\documentclass[conference,compsoc]{ieeeconf}  % Comment this line out if you need a4paper

\IEEEoverridecommandlockouts                              % This command is only needed if 
\usepackage{cite}
\usepackage{xcolor,soul}
\usepackage{tcolorbox} 
\soulregister\cite7
\soulregister\citep7
\soulregister\citet7
\soulregister\ref7
\soulregister\pageref7

\usepackage{amssymb}
\usepackage{bm}
\usepackage{flushend}

\usepackage{multirow}
\usepackage{soul}
\usepackage{svg}
\usepackage{float}

\usepackage{hyperref}

\hypersetup{
    colorlinks=true,
    linkcolor=black,
}

\usepackage{cuted}
\usepackage{capt-of}
\usepackage{caption}
\usepackage{amsmath}

\usepackage{blindtext}
\usepackage{tabularx}

\title{\LARGE \bf
PSS-BA: LiDAR Bundle Adjustment with Progressive Spatial Smoothing}

\author{Jianping Li, Thien-Minh~Nguyen, Shenghai Yuan, and Lihua Xie,~\IEEEmembership{Fellow,~IEEE}}

\begin{document}
% \begin{linenumbers}
% \pagewiselinenumbers 
% \switchlinenumbers
\maketitle

% \twocolumn[{%
% \renewcommand\twocolumn[1][]{#1}%
% \begin{center}
%     \centering
%     \captionsetup{type=figure}
%     \maketitle
%     \includegraphics[width=\textwidth]{fig/abstract.png}
%     \captionof{figure}{LiDAR Bundle Adjustment with Progressive Spatial Smoothing. Taking noisy LiDAR frames as input, PSS-BA progressively smooths out the noisy data and updates LiDAR poses, which ensures both the convergence and accuracy of state estimation. Ultimately, it enables the simultaneous generation of accurate LiDAR frames and parameterized surfaces and facilitates 3D modeling.}
%     \label{fig:abstract}
% \end{center}%
% }]

\renewcommand{\thefootnote}{}
\footnotetext{This research was conducted under project WP5 within the Delta-NTU Corporate Lab with funding support from A*STAR under its IAF-ICP programme (Grant no: I2201E0013) and Delta Electronics Inc. This research is also supported by the National Research Foundation, Singapore, under its Medium-Sized Center for Advanced Robotics Technology Innovation (CARTIN). All authors are with School of Electrical and Electronic Engineering, Nanyang Technological University, Singapore 639798, 50 Nanyang Avenue. (E-mail: jianping.li@ntu.edu.sg, thienminh.nguyen@ntu.edu.sg, shyuan@ntu.edu.sg, elhxie@ntu.edu.sg)}

%%%%%%%%%%%%%%%%%%%%%%%%%%%%%%%%%%%%%%%%%%%%%%%%%%%%%%%%%%%%%%%%%%%%%%%%%%%%%%%%

\begin{abstract}

    Accurate and consistent construction of point clouds from LiDAR scanning data is fundamental for 3D modeling applications. Current solutions, such as multiview point cloud registration and LiDAR bundle adjustment, predominantly depend on the local plane assumption, which may be inadequate in complex environments lacking of planar geometries or substantial initial pose errors. To mitigate this problem, this paper presents a LiDAR bundle adjustment with progressive spatial smoothing, which is suitable for complex environments and exhibits improved convergence capabilities. The proposed method consists of a spatial smoothing module and a pose adjustment module, which combines the benefits of local consistency and global accuracy. With the spatial smoothing module, we can obtain robust and rich surface constraints employing smoothing kernels across various scales. Then the pose adjustment module corrects all poses utilizing the novel surface constraints. Ultimately, the proposed method simultaneously achieves fine poses and parametric surfaces that can be directly employed for high-quality point cloud reconstruction. The effectiveness and robustness of our proposed approach have been validated on both simulation and real-world datasets. The experimental results demonstrate that the proposed method outperforms the existing methods and achieves better accuracy in complex environments with low planar structures.
    
\end{abstract}

%%%%%%%%%%%%%%%%%%%%%%%%%%%%%%%%%%%%%%%%%%%%%%%%%%%%%%%%%%%%%%%%%%%%%%%%%%%%%%%%
\section{Introduction }

Accurate 3D reconstruction of in GNSS-denied complex environment is a fundamental task in the fields of photogrammetry \cite{li2023whu,li2024hcto,liao2023se} and robotics \cite{xu2022fast,nguyen2023slict}, in which the exported point clouds is the basis for many tasks such as facility inspection \cite{cao2021distributed}, building information modeling (BIM) \cite{kim2021development}, and robot navigation \cite{liu2015robotic}. Although commercial mobile mapping systems (e.g., GeoSLAM, Leica Pegasus, etc.) equipped with expensive sensors have shown their powerful mapping ability over the past decade \cite{li2018automatic,li2020slam}, with recent developments in light-weight solid LiDAR technologies \cite{liu2021low}, research on improving the quality of point cloud collected by low-cost mapping systems is a hot topic in both academia and industry \cite{yang2022hierarchical,li2019nrli,deng2023long}.

In general, point cloud data obtained from a mobile mapping system are subjected to two sources of error: the pose error and the LiDAR measurement error. Most existing SLAM systems utilize some sequential registration scheme (e.g., ICP \cite{segal2009generalized}, NDT \cite{einhorn2015generic}, features \cite{zhang2014loam}) to merge the laser scans together, the incremental processing will cause the accumulated error in the pose estimates. The pose error can be corrected using some LiDAR bundle adjustment method \cite{liu2023efficient} (which we will review in Section \ref{review_ba}). As for the LiDAR measurement error, it is mainly caused by the range measurement noise, which can be corrected using point cloud smoothing or map-centric optimization \cite{park2021elasticity} in 3D space (a thorough review is given in Section \ref{review_pcs}). Intuitively, we found that the LiDAR bundle adjustment and point cloud smoothing \cite{han2017review} processes can benefit each other. On one hand, even with the imperfect initial poses, the local shape prior of the noisy point clouds can still be recognized by applying smoothing kernels \cite{han2017review} in 3D space. The shape prior can be used as a more general factor in bundle adjustment than the planar features. On the other hand, the adjusted poses guarantee the global accuracy of the point clouds compared to only conducting point cloud smoothing in a local 3D space. The related works of LiDAR bundle adjustment and point cloud smoothing are reviewed as follows.

% point cloud accuracy improvement in two aspects: LiDAR Bundle Adjustment, Point Cloud Smoothing
\subsection{LiDAR Bundle Adjustment}\label{review_ba}

% image bundle adjustment
Bundle adjustment, a fundamental technique originating from photogrammetry, aims to simultaneously estimate sensor poses and 3D feature coordinates in world frames \cite{triggs2000bundle}. This process is facilitated by extracting 2D point correspondences through image point features, hence enabling bundle adjustment factors via reprojection error. Conversely, in LiDAR bundle adjustment, the primary objective remains the estimation of all sensor poses. However, the inherent sparsity of point clouds in LiDAR data presents significant challenges in identifying precise point-level correspondences between frames. A concise and simple point-to-point correspondence is usually very hard to construct for pose correction in the LiDAR bundle adjustment task \cite{li2019nrli}. Therefore, the core issue of LiDAR bundle adjustment lies in the proper design and extraction of correspondences and factors.
% multiview registration 

LiDAR bundle adjustment can be regarded as an extension of pair-wise point cloud registration. Some works \cite{borrmann2008globally,koide2020interactive} conducted pair-wise registration \cite{koide2021voxelized} between all scanning data with overlaps, then the relative transformations are synchronized using a pose graph with outlier rejection \cite{xie2017graphtinker}. These methods work well for dense point clouds collected by terrestrial laser scanner \cite{dong2018hierarchical} and close-range structure-light scanner \cite{fantoni2012accurate}. However, the sparse LiDAR frames collected by light-weight solid-state laser scanners face great challenges obtaining accurate registration using existing point cloud registration algorithms \cite{segal2009generalized}. Moreover, the pose graph only considers relative transforms from pair-wise registration, loses information from raw point clouds, and restricts the mapping accuracies.

% ba in lidar slam

Recently, a number of studies \cite{liu2023efficient,zhou2021lidar} in the field of robotics have investigated the extraction of planar features from initial noisy point clouds, followed by conducting bundle adjustment utilizing these planar constraints. These methods have been successfully implemented in SLAM and sensor calibration \cite{liu2022targetless}. However, their reliance on planar features can lead to performance degradation and divergence in complex environments that lack structural features. Consequently, the applicability of existing planar feature-based LiDAR bundle adjustment methods is restricted, particularly in constructing BIM for complex buildings. Thus, this paper extracts more general feature correspondences using polynomial surface kernels with different scales to guarantee convergence and accuracy. Moreover, we analytically derive the jacobian of the polynomial residual with respect to the poses to speed up the optimization.

\subsection{Point Cloud Smoothing}\label{review_pcs}

Point cloud smoothing \cite{han2017review} and map-centric optimization \cite{park2021elasticity} improve the point cloud quality in the aspect of spatial consistency without considering the sensor poses, which are different from the existing LiDAR bundle adjustment that only optimizes the sensor poses. More specifically, the core assumption of most existing point cloud smoothing methods is the continuity and smoothness of the environments. The famous moving least squares (MLS) \cite{alexa2003computing} and its variations \cite{fleishman2005robust} are widely used for rendering and filtering of point clouds. MLS-based methods first determine the local shape using polynomial smooth kernels, then project the nearby points onto the fitted surface. With the fitting and projecting procedure, the point clouds are smoothed and the qualities are improved.

However, as the point cloud smoothing methods only maintain the local consistency of the point clouds, the pose drifts will not be corrected during the smoothing processing, and they can not guarantee global accuracy, especially for indoor mapping applications. Thus, this paper combines the point cloud smoothing with bundle adjustment, and also corrects the poses when conducting spatial smoothing to guarantee the global accuracy of the point clouds. Moreover, fixing the kernel size of existing point cloud smoothing methods may suffer from overfitting or underfitting of the local shape. This paper tries to progressively conduct spatial smoothing with different kernel sizes to achieve the convergence of the optimization.

Overall, to tackle the challenges of existing LiDAR bundle adjustment methods in complex environments, we propose PSS-BA. The core idea of the proposed PSS-BA is that, even with the imperfect initial poses, we still can obtain a shape prior by applying multi-scale smooth kernels on the initial noisy point clouds. Then the multi-scale smooth kernels are used for pose correction progressively. The main contributions of the proposed method can be stated as:

\begin{itemize}
\item PSS-BA utilizes the surface smooth kernel for constructing BA residuals providing more robust and richer constraints in complex environments compared to existing BA methods based on planar constraints.
\item PSS-BA introduces progressive smoothing accelerates BA convergence and improves accuracy compared to setting a fixed scale.
\item Ultimately, PSS-BA simultaneously achieves fine poses and parametric surfaces that can be directly employed for high-quality point cloud reconstruction and 3D modeling applications such as BIM.
\end{itemize}

The rest of this paper is structured as follows. A detailed description of the proposed method is presented in Section \ref{section_method}. The experiments are conducted on simulation and in-house datasets in Section \ref{section_exp}. Conclusion and future work are drawn in Section \ref{section_conclusion}.

\section{Methodology} \label{section_method}

\subsection{Notations and System Overview}

\subsubsection{Poses} In this paper, we use italic, bold lowercase, and bold uppercase letters to represent scalars, vectors, and matrices, respectively. Three main frames of reference are used in our proposed method, namely, the world frame $F^{W}$, the LiDAR frame $F^{L}$, and the tangent frame $F^{S}$ defined in a local 3D space \cite{dong2017novel}. We denote any point observed by the LiDAR in the $i$-th frame as $\mathbf{p}^{L}_i$. The pose for the $i$-th LiDAR frame is denoted as $[\mathbf{R}_i,\mathbf{t}_i]$, where $\mathbf{R}_i \in \mathrm{SO(3)}$ is the rotation matrix, $\mathbf{t}_i \in \mathbb{R}^3$ is the translation vector. All the poses are denoted as $\mathbf{X}=\{[\mathbf{R}_i,\mathbf{t}_i],i=0,1,...\}$.
The initial value is noted with breve $\breve{\circ}$, and the estimated value is noted with the hat $\hat{\circ}$.
The estimated pose for the $i^{th}$ LiDAR frame is denoted as $[\hat{\mathbf{R}}_i=\mathrm{Exp}{(\Delta \bm{\theta}_i)}\mathbf{R},\hat{\mathbf{t}}_i=\Delta \mathbf{t}_i+\mathbf{t}_i]$, where $\Delta\bm{\theta}_i \in \mathbb{R}^3 $ and $\Delta \mathbf{t}_i \in \mathbb{R}^3$ are the corresponding errors. The exponential map $\mathrm{Exp}:\mathbb{R}^3 \rightarrow \mathrm{SO(3)}$ has the form:
\begin{equation}
    \mathrm{Exp}(\Delta\bm{\theta}_i) = \mathbf{I} + \frac{\mathrm{sin}(|\Delta\bm{\theta}_i|)}{|\Delta\bm{\theta}_i|}[\Delta\bm{\theta}_i]_{\times} + \frac{1-\mathrm{cos}(|\Delta\bm{\theta}_i|)}{|\Delta\bm{\theta}_i|^{2}}[\Delta\bm{\theta}_i]^{2}_{\times}. \label{eq:Exp}
\end{equation}
The point $\hat{\mathbf{p}}_{i}^{W}$ in the world frame $F^{W}$ could be obtained using \eqref{eq:projection}.

\begin{equation}
    \hat{\mathbf{p}}_{i}^{W}=\hat{\mathbf{R}}_i\mathbf{p}_{i}^{L}+\hat{\mathbf{t}}_i, \label{eq:projection}
\end{equation}

\subsubsection{3D Normal} For the initial inaccurate normal $\hat{\mathbf{n}}_i$ associated with $\hat{\mathbf{p}}_{i}^{W}$, it satisfies $ |\hat{\mathbf{n}}|=1 $. Thus, the normal vector's degree of freedom is two, and it can be rewritten as:
\begin{equation}
    \hat{\mathbf{n}}_i \approx \mathbf{n}_i + [\hat{\mathbf{n}}^0_i,\hat{\mathbf{n}}^1_i]\Delta \bm{\phi}_i, \label{eq:normal_error}
\end{equation}
where $\Delta \bm{\phi}_i$ with shape of $2 \times 1$ represents the small errors. The vectors $\hat{\mathbf{n}}^0_i,\hat{\mathbf{n}}^1_i$ are two unit vectors and orthogonal to both $\hat{\mathbf{n}}_i$ and each other.

\subsubsection{Polynomial Smoothing Kernel} For a second-order polynomial surface defined within a local tangent space $F^{S}$, its functional form is as follows:

\begin{equation}
    z = f\left( x, y \right) = \bm{\alpha}^{\top}\left[ x^2, y^2,xy,x,y \right] ^{\top}, \label{eq:polynomial}
\end{equation}
where the vector $\bm{\alpha}$ with shape of $5 \times 1$ represents the coefficients that describe the surface's shape. Assuming a continuous 3D environment, the point clouds within the local tangent space are projected onto the surface $f$ to mitigate measurement noise.

\subsubsection{System Overview}

\begin{figure}[]
    \centering
    \includegraphics[width=0.48\textwidth]{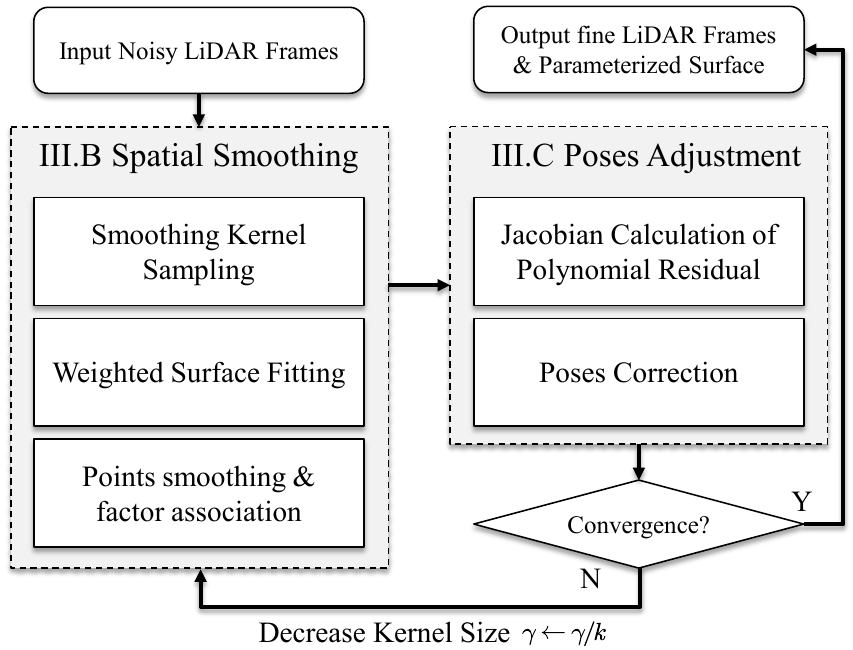}
    \caption{System overview of LiDAR bundle adjustment with progressive spatial smoothing (PSS-BA). }
    \label{fig:workflow}
\end{figure}

The proposed PSS-BA includes two key modules, namely spatial smoothing (\ref{sec:sptialsmooth}) and poses adjustment (\ref{sec:posescorrection}), which are illustrated in Fig.\ref{fig:workflow}. Taking the noisy LiDAR frames as input and a preset large smoothing kernel width $\gamma$, the spatial smoothing module smooths the noisy point cloud at a coarse scale and obtains the polynomial coefficients. Then the pose adjustment module utilizes the extracted polynomial coefficients to correct the lidar frames. If the change of correction is smaller than a threshold, it could be regarded as convergence. Otherwise, the smoothing kernel width $\gamma$ is decreased to a finer scale $\gamma \leftarrow \gamma/k$ to conduct the spatial smoothing again. The iterative smoothing and adjustment strategy guarantees the convergence and accuracy of the LiDAR bundle adjustment.

\begin{figure}
    \centering
    \includegraphics[width=0.48\textwidth]{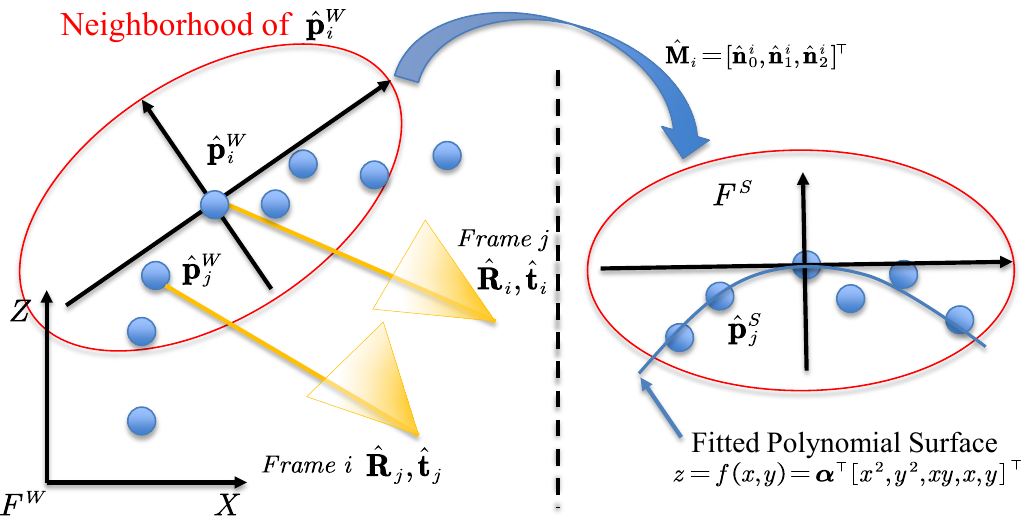}
    \caption{Surface fitting within a smoothing kernel. }
    \label{fig:surface_fitting}
\end{figure}

\begin{figure}
    \centering
    \includegraphics[width=0.48\textwidth]{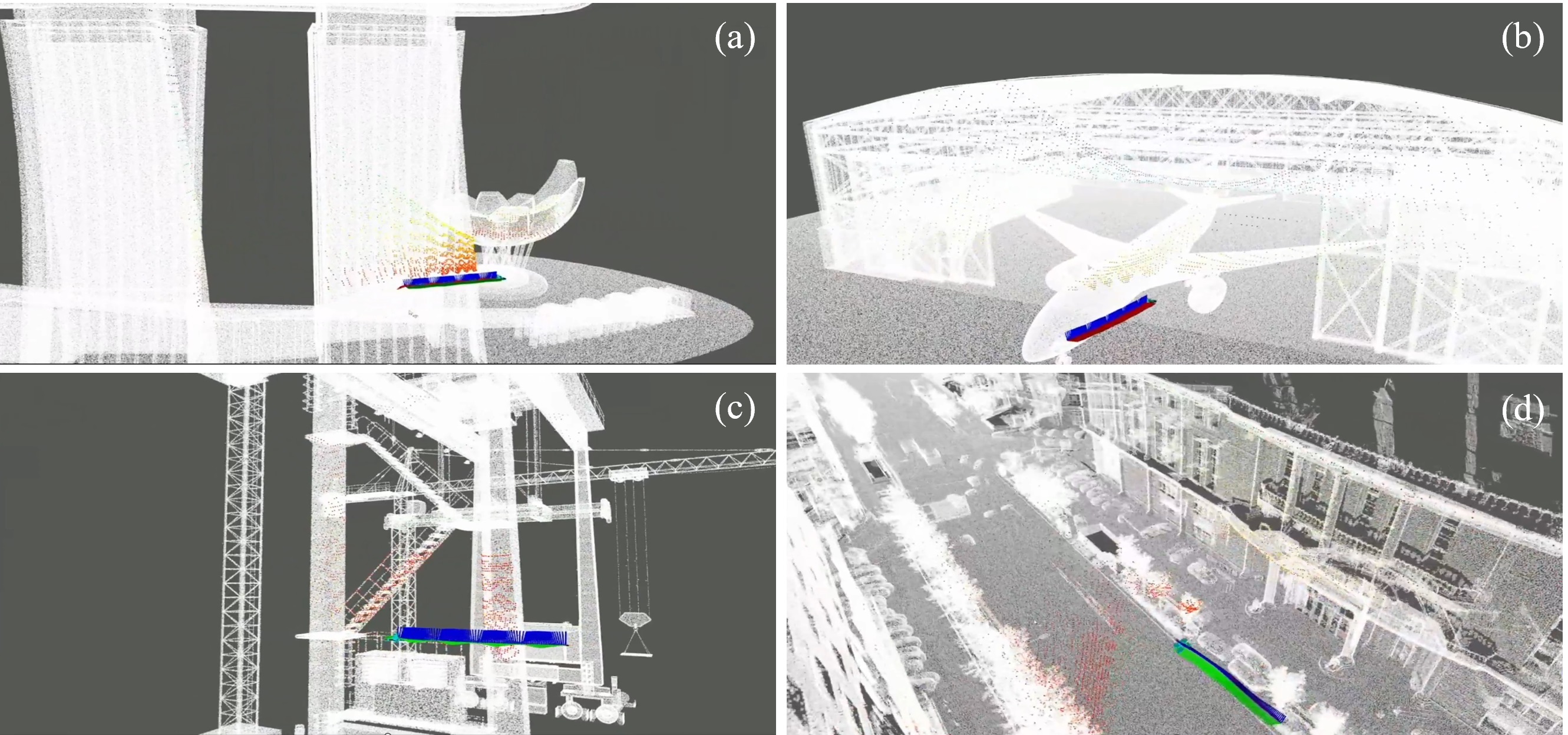}
    \caption{Simulation of UAV LiDAR sensing in complex environments using MARSIM \cite{kong2023marsim}. The colorful points belong to the current LiDAR frame. The white points are the ground truth point clouds. The axis sequences are the simulated trajectories of the UAV. (a) Singapore Marina Bay Sands (MBS). (b) Hangar. (c) Crane. (d) Street. }
    \label{fig:simulation}
    \vspace{-0.5cm}
\end{figure}
\subsection{Spatial Smoothing for LiDAR Factor Extraction \label{sec:sptialsmooth}}

\subsubsection{Smoothing Kernel Sampling} 
The input noisy point clouds are uniformly sampled using the voxel size of $\gamma$. Afterwards, the remaining points, $\{\hat{\mathbf{p}}^{W}_i, i \in \rm{\Psi}\}$, are treated as smoothing kernels. The initial point normals for each point in the noisy point clouds are calculated using Principal Component Analysis (PCA). For a given smoothing kernel, $\hat{\mathbf{p}}^{W}_i$, we define its neighboring points within the distance of $\gamma$ as $\{\hat{\mathbf{p}}^{W}_j,j\in\rm{\Phi}_i\}$. The inital point normals for the smoothing kernel $\hat{\mathbf{p}}^{W}_i$ and a neighborhood $\hat{\mathbf{p}}^{W}_j$ are respectively defined as $\breve{\mathbf{n}}_i$ and $\breve{\mathbf{n}}_j$.

Under the assumption of a continuous 3D environment, there is minimal variation in the normals across a local space. Consequently, we can obtain an optimal estimation of the normal, denoted as $\hat{\mathbf{n}}_i$ for the smoothing kernel $\hat{\mathbf{p}}^{W}_i$ by solving the objective function $\mathbf{G}(\hat{\mathbf{n}}_i)$ that constrains the change of normals respect to neighborhood normals:
\begin{equation}
    \begin{split}
    \underset{\hat{\mathbf{n}}_i,|\hat{\mathbf{n}}_i|=1}{\mathrm{argmin}}\ \mathbf{G}(\hat{\mathbf{n}}_i) &= 1-\hat{\mathbf{n}}^{\top}_i\breve{\mathbf{n}}_i + \mu |\mathbf{D}(\hat{\mathbf{n}}_i)|_{0},\\
    \mathbf{D}(\hat{\mathbf{n}}_i)_j&= 1-\hat{\mathbf{n}}^{\top}_i\breve{\mathbf{n}}_j,
    \end{split}\label{eq:normal}
\end{equation}
where $\mathbf{D}(\hat{\mathbf{n}}_i)$ is the differential function for $\hat{\mathbf{n}}_i$ in the 3D space. The $L_0$ normalization counting the non-zero item is used here to eliminate the influence of outliers and preserve the original shape in the sharp regions \cite{xu2011image,sun2015denoising}. $\mu$ is the weight that balances the data term and smooth term. Equation \eqref{eq:normal} can be minimized using an auxiliary function and is detailed in the appendix.

Once we have obtained the optimal normal $\hat{\mathbf{n}}_i$ for the smoothing kernel $\hat{\mathbf{p}}^{W}_i$, the local tangent frame $F^S$ is established for $\hat{\mathbf{p}}^{W}_i$ by constructing three orthogonal axises:
\begin{equation}
\hat{\mathbf{n}}^i_2 = \hat{\mathbf{n}}_i, \hat{\mathbf{n}}^i_1 = [\hat{{n}}_{i,1},-\hat{{n}}_{i,0},0]^{\top}, \hat{\mathbf{n}}^i_0 = \hat{\mathbf{n}}^i_1 \times \hat{\mathbf{n}}_i.\\
\end{equation}
With these three axises, $\hat{\mathbf{M}}_i = [\hat{\mathbf{n}}^i_0,\hat{\mathbf{n}}^i_1,\hat{\mathbf{n}}^i_2]^{\top}$ is the matrix that tranform the points from $F^W$ to $F^S$ as shown in Fig. \ref{fig:surface_fitting}.

\subsubsection{Weighted Surface Fitting}

For the smoothing kernel $\hat{\mathbf{p}}^{W}_i$, its neighborhood points $\{\hat{\mathbf{p}}^{W}_j,j\in\rm{\Phi}_i\}$ are projected to smoothing kernel's tangent space $F^S$:
\begin{equation}
    \left[ \hat{x}_{j}^{S},\hat{y}_{j}^{S},\hat{z}_{j}^{S} \right] ^{\top} =\hat{\mathbf{p}}_{j}^{S}=\hat{\mathbf{M}}_i\left( \hat{\mathbf{p}}_{j}^{W}-\hat{\mathbf{p}}_{i}^{W} \right).
\end{equation}

Let the paramters for the smoothing kernel $\hat{\mathbf{p}}^{W}_i$ be $\bm \alpha^i = [\alpha^i_0, \alpha^i_1,...,\alpha^i_5]^\top$, the smoothed coordinates of $\hat{\mathbf{p}}_{j}^{S}$ is $[\hat{x}_{j}^{S},\hat{y}_{j}^{S},f_i\left( \hat{x}_{j}^{S},\hat{y}_{j}^{S} \right)]^\top$, where $f_i\left( \hat{x}_{j}^{S},\hat{y}_{j}^{S} \right)$ is the polynomial function and calculated as follow:
\begin{equation}
    f_i\left( \hat{x}_{j}^{S},\hat{y}_{j}^{S} \right) = {\bm{\alpha}^i}^{\top}\left[ \left( \hat{x}_{j}^{S} \right) ^2,\left( \hat{y}_{j}^{S} \right) ^2,\hat{x}_{j}^{S}\hat{y}_{j}^{S},\hat{x}_{j}^{S},\hat{y}_{j}^{S} \right]^{\top}.
\end{equation}
Now the core problem is to robustly determine the optimal parameter $\bm \alpha^i$ using the projected noisy points $\{\hat{\mathbf{p}}^{S}_j,j\in\rm{\Phi}_i\}$. We find the best parameters $\bm \alpha^i$ using least-square estimation considering the Gaussian radial weight function $\mathrm{w}(d)$\cite{alexa2003computing}:  
\begin{equation}
    \begin{split}
    \underset{\bm \alpha^i}{\mathrm{argmin}}\ \Sigma_{j\in\rm{\Phi}_i}& \left|\big(f_i\left( \hat{x}_{j}^{S},\hat{y}_{j}^{S} \right) - \hat{z}_{j}^{S}\big)\mathrm{w}(|\hat{\mathbf{p}}_{j}^{S}|)\right|^2,\\
    &\mathrm{w}(d)= e^{-d^2/\gamma^2},
    \end{split}
\end{equation}
Then the fitted polynomial surface is obtained as shown in Fig. \ref{fig:surface_fitting}.

\subsubsection{Points Smoothing and Factor Association}
By replacing $\hat{z}_{j}^{S}$ with $f_i\left( \hat{x}_{j}^{S},\hat{y}_{j}^{S} \right)$ for each point in $\{\hat{\mathbf{p}}^{S}_j,j\in\rm{\Phi}_i\}$, we could obtain the smoothed point clouds. Moreover, the difference between $\hat{z}_{j}^{S}$ and $f_i\left( \hat{x}_{j}^{S},\hat{y}_{j}^{S} \right)$ is regarded as the error caused by the poses' error. Thus we associate $\{\hat{\mathbf{p}}^{S}_j,j\in\rm{\Phi}_i\}$ with $\mathbf{p}^{S}_i$, and use it for the following poses adjustment if the proposed PSS-BA still has not converged.
\subsection{Poses Adjustment using Polynomial Surface Constraints \label{sec:posescorrection}}

\subsubsection{Jacobian calculation of polynomial residual}
The difference between $\hat{z}_{j}^{S}$ and the fitted polynomial surface $f_i\left( \hat{x}_{j}^{S},\hat{y}_{j}^{S} \right)$ is regarded as the error $\sigma _{i,j}$ and written as follow:
\begin{equation}
    \sigma _{i,j}=f_i\left( \hat{x}_{j}^{S},\hat{y}_{j}^{S} \right) -\hat{z}_{j}^{S}.
\end{equation}
$\sigma _{i,j}$ is correlated with $i^{th}$ and $j^{th}$ poses, namely $[\hat{R}_i,\hat{t}_i]$ and $[\hat{R}_j,\hat{t}_j]$ as shown in Fig. \ref{fig:surface_fitting}. The jacobian of $\sigma _{i,j}$ with respect to the $i^{th}$ pose can be derived using the chain rule.

\subsubsection{Poses correction} 
Finally, we combine all the polynomial residuals to correct the poses using least-square estimation as follows:
\begin{equation}
    \begin{split}
        \underset{\hat{\mathbf{X}}}{\mathrm{argmin}}\ \Sigma_{i \in \Psi} \Sigma_{j \in \rm{\Phi}_i} \left|\sigma_{i,j}\right|^2.
    \end{split}
\end{equation}
If the change of the norm for $\hat{\mathbf{X}}$ after optimization is smaller than a threshold $\mathrm{T}_{\mathrm{conv}}$, the proposed PSS-BA is terminated. Otherwise, PSS-BA will decrease the kernel size and continue the next iteration.

\section{Experiments} \label{section_exp}

\begin{figure}
    \centering
    \includegraphics[width=0.5\textwidth]{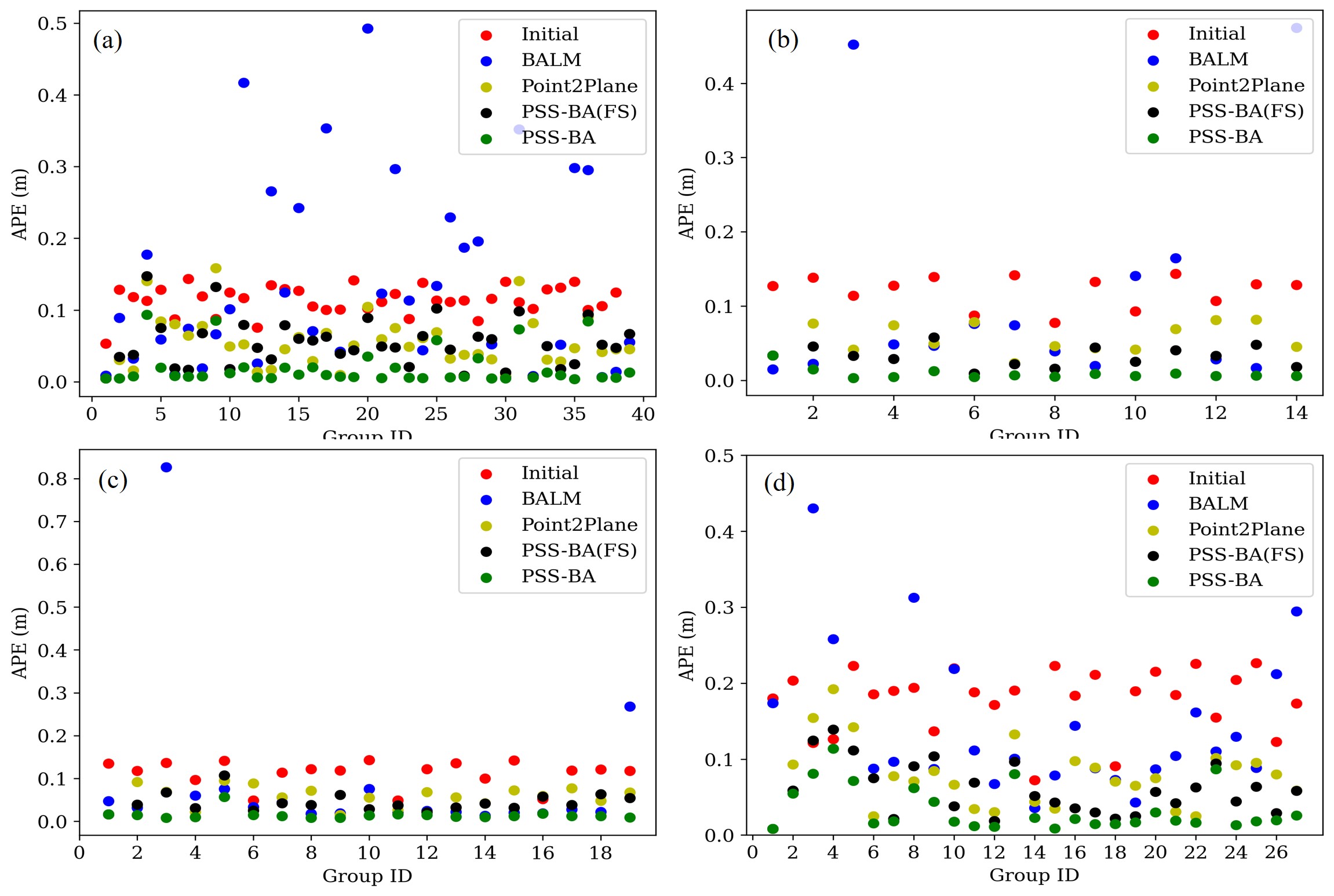}
    \caption{Evaluation of different methods on the simulation datasets. (a) Singapore Marina Bay Sands (MBS); (b) Hangar; (c) Crane; (d) Street. }
    \label{fig:evaluation}
\end{figure}

\begin{figure}
    \centering
    \includegraphics[width=0.5\textwidth]{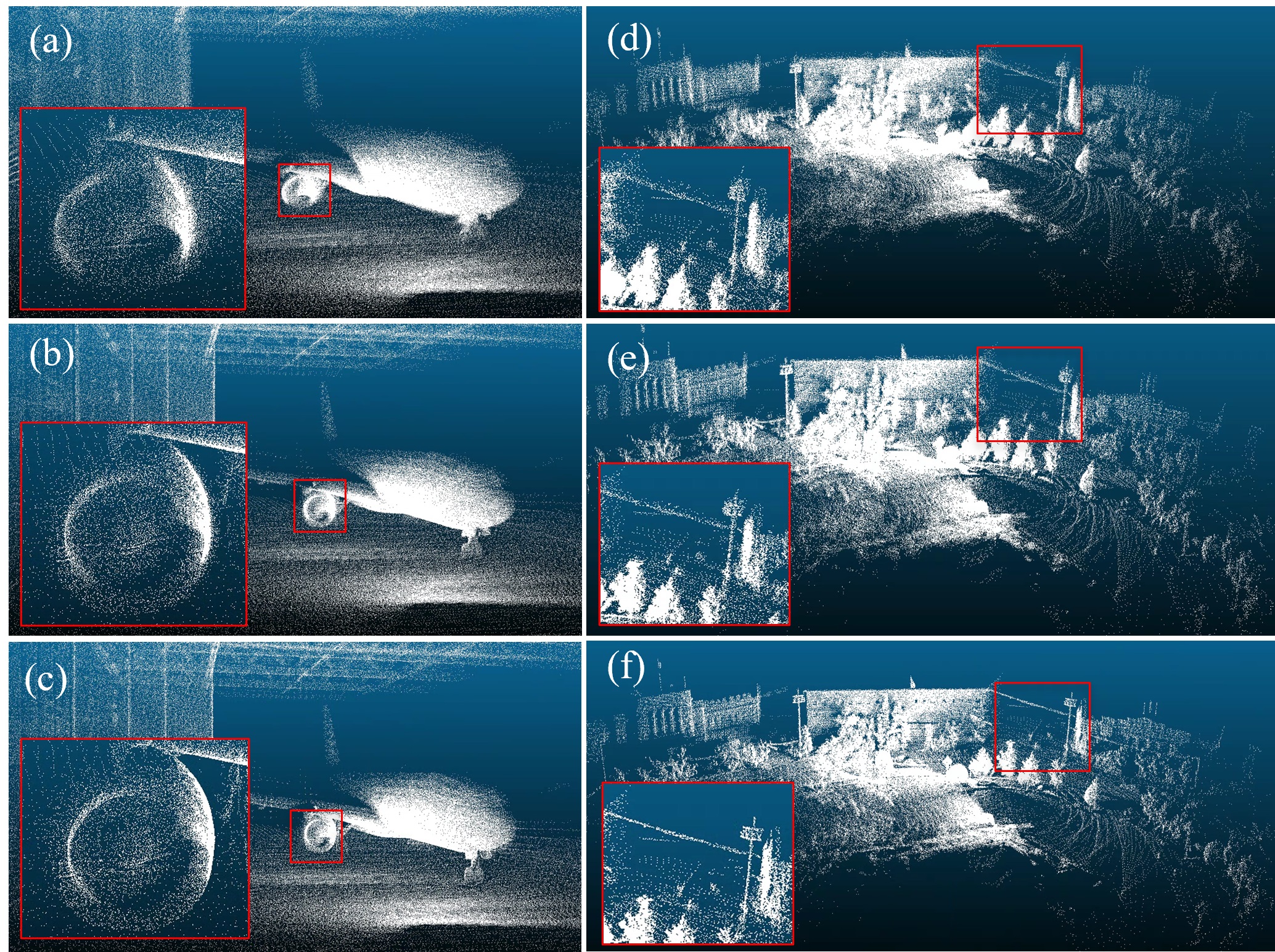}
    \caption{Visual comparison of different methods. (a),(b), and (c) are the results from initial values, BALM, and PSS-BA on the hangar dataset, respectively. (d),(e), and (f) are the results from initial values, BALM, and PSS-BA on the street dataset, respectively.}
    \label{fig:evaluation_vis}
\end{figure}

\subsection{Implementation}
We implemented the proposed PSS-BA in C++. The initial kernel size $\gamma$ is set to 3 m. The decreasing factor $k$ is set to 1.4. The balance factor $\mu$ for $L_0$ optimization of normal is set to 0.05. The termination threshold $\mathrm{T}_{\mathrm{conv}}$ is set to 0.01. The size of the total frames to be optimized is limited to 100 considering the high memory usage of LiDAR BA algorithms. All algorithms are evaluated on a computer with an Intel Core I9-I2900 CPU.

\subsection{Simulation Dataset}
To thoroughly evaluate the proposed method, we first simulate the laser scanning frames using MARSIM \cite{kong2023marsim} in various environments as shown in Fig. \ref{fig:simulation}. The laser sensor used for simulation is a low-cost LiDAR, Livox Mid360 \cite{livox2023mid}, which is now widely used for 3D mapping. The first three ground truth point clouds are provided by the CARIC \cite{caric2023} benchmark, which mainly focuses on inspection applications with irregular shapes and 3D curves. The last ground truth point clouds with 3-centimeter accuracy were collected using a mobile mapping system \cite{li2018automatic} in the urban environment. The UAV trajectories for simulation were pre-defined manually. Then the simulated laser scanning data are split into groups with 100 sequential frames. We added Gaussian noises to the ground truth poses (0.2 m and 1 deg for translation and rotation, respectively) to simulate the initial inaccurate values, which are used as the input for different LiDAR BA algorithms.

% describe the dataset

The SOTA method, BALM \cite{liu2023efficient}, which utilizes planar voxel as the adjustment constraints is compared. Furthermore, we modified the PSS-BA by (1) using only the point-to-plane constraints (Point2Plane) \cite{fantoni2012accurate}; (2) using only a fixed scale without changing the smoothing kernel width $\gamma$ (PSS-BA[FS]) as the ablation.  Absolute Position Error (APE) \cite{grupp2017evo} is used to evaluate the accuracies of different methods as plotted in Fig. \ref{fig:evaluation}. It could be found that BALM sometimes diverges because it only can not get enough constraints in the environments without enough planar features. The average APE is listed in Table \ref{tab:ate_simulation}. It should be noted that the divergence cases are removed to get meaningful statistics for BALM. Table \ref{tab:ate_simulation} demonstrates that the proposed PSS-BA outperforms other methods in the simulation datasets.

Unlike the Point2Plane strategy, which depends solely on planar features, the proposed PSS-BA yields superior results in environments with a lot of 3D curvatures. The polynomial surface residual used in the PSS-BA is more accurate than the local planar assumption for the complex environments. Furthermore, when compared with PSS-BA (FS), our progressive smoothing strategy offers improved outcomes by effectively balancing convergence and precision. Some visual comparisons of the resulting point clouds are illustrated in Fig. \ref{fig:evaluation_vis}, which demonstrates that the proposed PSS-BA could obtain sharp and accurate point clouds for the complex 3D shapes.

\begin{table}[h]
    \caption{ATE of PSS-BA and other methods on simulation dataset (Unit [m]). The best results are in \textbf{BOLD}, second best results are in \underline{underlined}.\label{tab:ate_simulation}}
    \centering
    \begin{tabularx}{0.48\textwidth}{cccccc}\hline\hline
    Dataset & Initial & BALM & Point2Plane & PSS-BA(FS) & PSS-BA \\ \hline
    MBS     &  0.11   &  0.09   &   \underline{0.05}    &    \underline{0.05}    &  \textbf{0.02}  \\
    Hangar  &  0.12     &  0.07   &    0.06      &   \underline{0.03}          &  \textbf{0.01}  \\
    Crane   &  0.11     &  0.07 &    0.06    &    \underline{0.04}    &  \textbf{0.01} \\ 
    Street  &  0.17     &  0.09  &    0.08     &    \underline{0.06}      &  \textbf{0.03} \\ \hline\hline
    \end{tabularx}
\end{table}

\subsection{Real-world Dataset}

\begin{figure}
    \centering
    \includegraphics[width=0.48\textwidth]{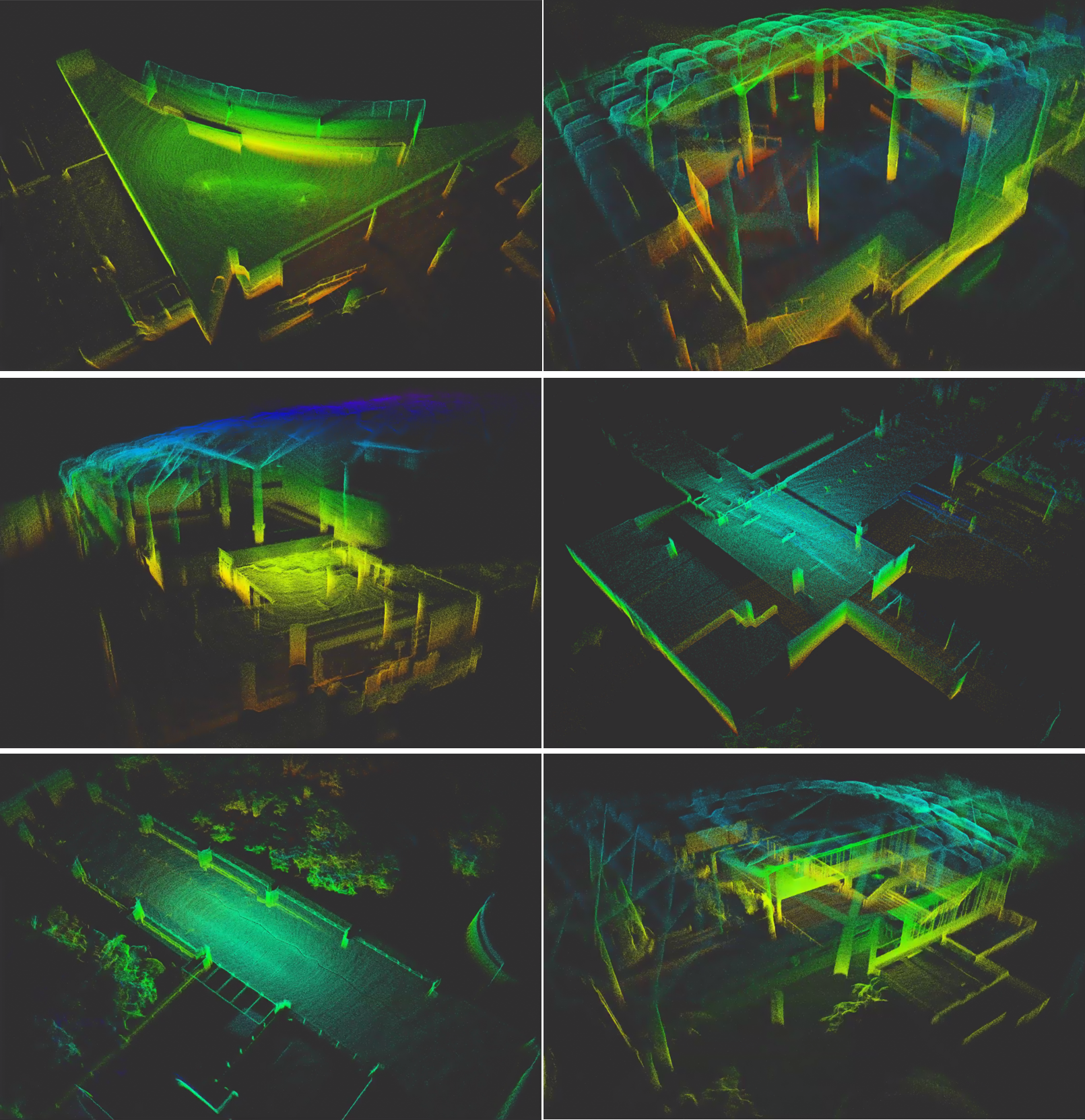}
    \caption{The sample point clouds resulting from PSS-BA in the Nanyang Technological University (NTU) campus.}
    \label{fig:real_evaluation_vis}
    \vspace{-0.5cm}
\end{figure}

We collected a lot of in-house datasets using a helmet-based mapping system consisting of a Livox Mid360 laser scanner \cite{li2024hcto} in the Nanyang Technological University (NTU) campus. The initial values for the bundle adjustment are also obtained using Fast-LIO \cite{xu2022fast}. The laser scanning data are split into groups with 100 sequential frames. Some sample results from the proposed PSS-BA are shown in Fig. \ref{fig:real_evaluation_vis}. From the visual inspection, the resulting point clouds from the proposed PSS-BA are very sharp and clear, even the indoor environments with irregular and complex 3D shapes, which may cause difficulty for the plane-based bundle adjustment methods.

% \begin{figure}
%     \centering
%     \includegraphics[width=0.48\textwidth]{fig/hardware.png}
%     \caption{An in-house helmet-based mapping system.}
%     \label{fig:hardware}
% \end{figure}

\begin{figure}
    \centering
    \includegraphics[width=0.4\textwidth]{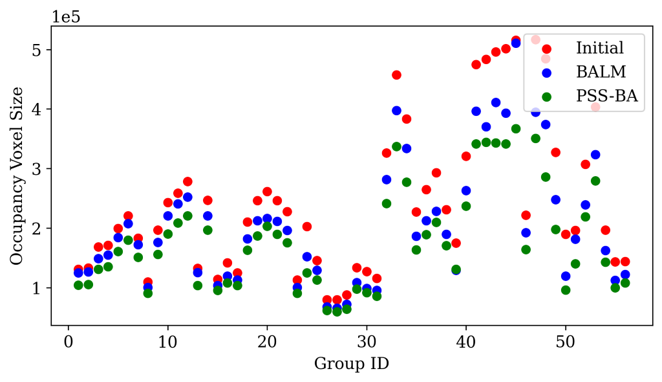}
    \caption{Evaluation of different methods on the real dataset. The average occupancy voxel sizes for initial values, BALM, and PSS-BA are 241717, 203940, and 176480, respectively.}
    \label{fig:real_evaluation}
    \vspace{-0.5cm}
\end{figure}

To evaluate the mapping accuracies, we adopted the method proposed by \cite{filatov20172d} and commonly used for point cloud accuracies without ground truth \cite{liu2023efficient}. The evaluation criteria are intuitive, and count the occupancy of resulting point clouds. If the point clouds are registered with each other very well, they should occupy less voxels. Based on this assumption, we divided the 3D space into voxels with the size of 0.1 m, then counted the voxels occupied and plotted in Fig. \ref{fig:real_evaluation}. As the Fast-LIO provided good initial poses, the occupancy voxel sizes are not improved a lot in some groups of the real-world datasets using the LiDAR bundle adjustment. However, as the operator moved very aggressively in the rest of the groups, the Fast-LIO may suffer from drifts. The proposed method could achieve better accuracies than the BALM, which only relies on planar features. Overall, the average occupancy voxel sizes for initial values, BALM, and PSS-BA are 241717, 203940, and 176480, respectively. The statistics indicate that the proposed PSS-BA could achieve the best mapping accuracies in complex indoor environments.

\section{Conclusion} \label{section_conclusion}

This paper proposes PSS-BA, targeting the accurate construction of point clouds from LiDAR data in complex environments. Existing solutions predominantly depend on detecting planar features, which may be inadequate in complex environments with less structural geometry or substantial initial pose errors. PSS-BA overcomes these challenges by combining spatial smoothing and pose adjustment modules, ensuring local consistency and global accuracy, leading to precise poses and high-quality point cloud reconstruction.
PSS-BA's effectiveness has been demonstrated in simulation and real-world datasets, offering a promising solution for accurate 3D modeling in challenging environments.

% Due to the memory limitation, PSS-BA can only process several hundred frames together, which limits the PSS-BA's application on large-scale mapping applications. In the near future, we will extend PSS-BA to large-scale applications by seeking hierarchical adjustment strategies.

\bibliographystyle{ieeetr}
\bibliography{ref}

% \end{linenumbers}

\addtolength{\textheight}{-2cm}  

\end{document}